\documentclass{article} 
\usepackage{nips14submit_e,times}
\usepackage{hyperref}
\usepackage{url}
\usepackage{amsmath}
\usepackage{algorithm}
\usepackage{algorithmic}

\newenvironment{singlespacedlist}{
\begin{itemize}{}{}
	\setlength{\itemsep}{0pt}
	\setlength{\parskip}{0pt}
	\setlength{\parsep}{0pt}}{\end{itemize}
}

\title{A Hybrid Latent Variable Neural Network Model for Item Recommendation}

\author{
Michael R. Smith\\
Department of Computer Science\\
Brigham Young University\\
Provo, UT 86402 \\
\texttt{msmith@axon.cs.byu.edu} \\
\And
Tony Martinez \\
Department of Computer Science \\
Brigham Young University \\
Provo, UT 84602 \\
\texttt{martinez@cs.byu.edu} \\
\And
Michael Gashler \\
Department of Computer Science and Computer Engineering \\
University of Arkansas \\
Fayetteville, AR 72701 \\
\texttt{mgashler@uark.edu} \\
}

\nipsfinalcopy 

\begin{document}

\maketitle

\begin{abstract}
Collaborative filtering is used to recommend items to a user without requiring a knowledge of the item itself and tends to outperform other techniques.
However, collaborative filtering suffers from the \textit{cold-start} problem, which occurs when an item has not yet been rated or a user has not rated any items.
Incorporating additional information, such as item or user descriptions, into collaborative filtering can address the cold-start problem.
In this paper, we present a neural network model with latent input variables (\textit{latent neural network} or LNN) as a hybrid collaborative filtering technique that addresses the cold-start problem.
LNN outperforms a broad selection of content-based filters (which make recommendations based on item descriptions) and other hybrid approaches while maintaining the accuracy of state-of-the-art collaborative filtering techniques.

\end{abstract}

\section{Introduction}
Modern technology enables users to access an abundance of information.
This deluge of data makes it difficult to sift through it all to find what is desired.
This problem is of particular concern to companies who are trying sell products (e.g. Amazon or Walmart) or recommend movies (e.g. Netflix).
To lessen the severity of information overload, recommender systems help a user find what he or she is looking for. 
Two commonly used classes of recommender systems are content-based filters and collaborative filters.

Content-based filters (CBF) make recommendations based on item/user descriptions and users' ratings of the items.
Creating item/user descriptions that are predictive of how a user will rate an item, however, is not a trivial process.
On the other hand, collaborative filtering (CF) techniques use correlations between users' ratings to infer the rating of unrated items for a user and make recommendations without having to understand the item or user itself.
CF does not depend on item descriptions and tends to produce higher accuracies than CBF.
However, CF suffers from the \textit{cold-start problem} which occurs when an item cannot be recommended unless it is has been rated before (\textit{first-rater problem}) or when a user has not rated any items (\textit{new-user problem}).
This is particularly important in domains where new items are frequently added to a set of items and users are more interested in the new items.
For example, many users are more interested, and likely to purchase, new styles of shoes rather than out-dated styles or many users are more interested in watching newly released movies rather than older movies.
Recommending old items has the potential to drive away customers.
In addition, making inappropriate recommendations for new users who have not built a profile can also drive away users.

One approach for addressing the cold-start problem is using a hybrid recommender system that can leverage the advantages of multiple recommendation systems.
Developing hybrid models is a significant research direction \cite{Claypool1999,Schein2002,Li2003,Su2007,Cremonesi2011,Forbes2011,Lin2013}.
Many hybrid approaches combine a content-based filter with a collaborative filter through methods such as averaging the predicted ratings or combining the top recommendations from both techniques \cite{Burke2002}.
In this paper, we present a neural network model with latent input variables (\textit{latent neural network} or LNN) as a hybrid recommendation algorithm that addresses the cold-start problem.
LNN uses a matrix of item ratings and item/user descriptions to simultaneously train the weights in a neural network and induce a set of latent input variables for matrix factorization.
Using a neural network allows for flexible architecture configurations to model higher-order dependencies in the data.

LNN is based on the idea of generative backpropagation (GenBP) \cite{hinton:generative_backpropagation} and expands upon unsupervised backpropagation (UBP) \cite{Gashler2014}.
Both GenBP and UBP are neural network methods that induce a set of latent input variables.
The latent input variables form an internal representation of observed values.
When the latent input variables are fewer than the observed variables, both methods are dimensionality reduction techniques.
GenBP adjusts its latent inputs while holding the network weights constant.
It has been used to generate labels for images \cite{coheh:label_images_generative_backprop}, and for natural language \cite{bengio:neural_language_generative_backprop}.
UBP differs from GenBP in that it trains network weights simultaneously with the latent inputs, instead of training the weights as a pre-processing step.
LNN is a further development of UBP that incorporates input features among the latent input variables.
By incorporating user/item descriptions as input features, LNN is able to address the cold-start problem.
We find that LNN outperforms other content-based filters and hybrid filters on the cold-start problem.
Additionally, LNN outperforms its predecessor (UBP) and maintains an accuracy similar to matrix factorization (which cannot handle the cold-start problem) on non-cold-start recommendations.


\section{Related Work}
\label{section:relatedwork}
Matrix factorization (MF) has become a popular technique, in part due to its effectiveness with the data used in the NetFlix competition \cite{koren:matrix_factorization, Salakhutdinov2007} and is widely considered a state-of-the-art recommendation technique.
MF is a linear dimensionality reduction technique that factors the rating matrix into two much-smaller matrices.
These smaller matrices can then be combined to predict all of the missing ratings in the original matrix.
It was previously shown that MF could be represented with a neural network model involving one hidden layer and linear activation functions \cite{takacs2009scalable}.
By using non-linear activation functions, unsupervised backpropagation (UBP) may be viewed as a non-linear generalization of MF.
UBP is related to nonlinear PCA (NLPCA) that was used as a means of imputing missing values (a task similar to recommending items) \cite{nlpca}.
UBP utilizes three phases for training to initialize the latent variables, the weights of the model and then to update the weights and latent variables simultaneously.
LNN further builds on UBP and NLPCA by integrating item or user descriptions with the latent input variables.

Pure collaborative filtering (CF) techniques are not able to handle the cold-start problem for items or users.
As a result, several hybrid methods have been developed that incorporate item and/or user descriptions into collaborative filtering approaches.
The most common, as surveyed by Burke \cite{Burke2002}, involves using separate CBF and CF techniques and then combining their outputs (i.e. weighted average, combining the output from both techniques, or switching depending on the context) or using the output from one technique as input to another.
Content-boosted collaborative filtering \cite{Melville2004} uses CBF to fill in the missing values in the ratings matrix and then the dense ratings matrix is passed to a collaborative filtering method (in their implementation, a neighbor based CF).
Other work addresses the cold-start problem by build user/item descriptions for later use in a recommendation system \cite{Zhou2011}.

\section{Latent Neural Network}
\label{section:LNN}
In this section, we formally describe \textit{latent neural networks} (LNN).
At a high-level, a LNN is a neural network with latent input variables induced using generative backpropagation.
Put simply, generative backpropagation calculates the gradient of the latent inputs with respect to the error and updates them in a manner similar to how the weights are updated in the backpropagation algorithm.

\subsection{Preliminaries}
In order to formally describe LNNs, we define the following terms.

\begin{singlespacedlist}
	\item Let $\mathbf{X}$ be a given $m\times n$ sparse user/item rating matrix, where $m$ is the number of items and $n$ is the number of users.
	\item Let $\mathbf{A}$ be an $m\times a$ matrix, representing the given portion of the item profiles.
	\item Let $\mathbf{V}$ be an $m\times t$ matrix, representing the latent portion of the item profiles.
	\item If $x_{rc}$ is the rating for item $r$ by user $c$ in $\mathbf{X}$, then $\hat{x}_{rc}$ is the predicted rating when $\mathbf{a}_r\in\mathbf{A}$ and $\mathbf{v}_r\in\mathbf{V}$ are concatenated into a single vector $\mathbf{q}_r$ and then fed forward into the LNN.
	\item Let $w_{ij}$ be the weight that feeds from unit $i$ to unit $j$ in the LNN.
	\item For each network unit $i$ on hidden layer $j$, let $\beta_{ji}$ be the net input into the unit, $\alpha_{ji}$ be the output or activation value of the unit, and $\delta_{ji}$ be an error term associated with the unit.
	\item Let $l$ be the number of hidden layers in the LNN.
	\item Let $\mathbf{g}$ be a vector representing the gradient with respect to the weights of the LNN, such that $g_{ij}$ is the component of the gradient that is used to refine $w_{ij}$.
	\item Let $\mathbf{h}$ be a vector representing the gradient with respect to the latent inputs of the LNN, such that $h_i$ is the component of the gradient that is used to refine $v_{ri}\in\mathbf{v}_r$.
\end{singlespacedlist}
We use item descriptions, but user descriptions could easily be used by transposing the $\mathbf{X}$ and using user descriptions instead of item descriptions.

As using generative backpropagation to compute the gradient with respect to the latent inputs, $\mathbf{h}$, is less commonly used, we provide a derivation of it here.
We compute each $h_i\in\mathbf{h}$ from the presentation of a single element $x_{rc}\in\mathbf{X}$ 
since we assume that $\mathbf{X}$ is typically high-dimensional and sparse.
It is significantly more efficient to train with the presentation of each known element individually.
We begin by defining an error signal for an individual element, $E_{rc}=(x_{rc}-\hat{x}_{rc})^2$, and then express the gradient as the partial derivative of this error signal with respect to each latent input (the non-latent inputs in $\mathbf{A}$ do not change): 
\begin{equation}
	h_i=\frac{\partial E_{rc}}{\partial v_{ri}}.
	\label{eq_begin}
\end{equation}
The intrinsic input $v_{ri}$ affects the value of $E_{rc}$ through the net value of a unit ($\beta_{ji}$) and further through the output of a unit ($\alpha_{ji}$).
Using the chain rule, Equation \ref{eq_begin} becomes:
\begin{equation}
	h_i=\frac{\partial E_{rc}}{\partial \alpha_{0c}}\frac{\partial \alpha_{0c}}{\partial \beta_{0c}}\frac{\partial \beta_{0c}}{\partial v_{ri}}
\end{equation}
where $\alpha_{0c}$ and $\beta_{0c}$ represent, respectively, the output values and the net input values of the output nodes (the $0^{th}$ layer).
The backpropagation algorithm calculates $\frac{\partial E_{rc}}{\partial \alpha_{0c}}\frac{\partial \alpha_{0c}}{\partial \beta_{0c}}$ (which is $\frac{\partial E_{rc}}{\partial \beta_{j,i}}$ for a network unit) as the error term $\delta_{ji}$ associated with a network unit.
Thus, to calculate $h_i$, the only additional calculation to the backpropagation algorithm that needs to be made is $\frac{\partial \beta_{jc}}{\partial v_{ri}}$.
For a single layer perceptron (0 hidden layers):
$$\frac{\partial \beta_{0c}}{\partial v_{ri}} = \frac{\partial}{\partial v_{ri}} \sum_t w_{tc}\: v_{rt}$$ 
which is non-zero only when $t$ equals $i$ and is equal to $w_{ic}$ since the error is being calculated with respect to a single element in $\mathbf{X}$.
When there are no hidden layers ($l=0$) and using the error from a single element $x_{rc}$:
\begin{equation}
	h_i=-w_{ic} \delta_c.
	\label{eq_no_hidden}
\end{equation}
If there is at least one hidden layer ($l > 0$), then,
$$\frac{\partial \beta_{0c}}{\partial v_{ri}} = \frac{\partial \beta_{0c}}{\partial\boldsymbol{\alpha}_1} \frac{\partial \boldsymbol{\alpha}_1}{\partial \boldsymbol{\beta}_1}\dots\frac{\partial \boldsymbol{\alpha}_l}{\partial \boldsymbol{\beta}_l}\frac{\partial \boldsymbol{\beta}_l}{\partial v_{ri}},$$
where $\boldsymbol{\alpha}_k$ and $\boldsymbol{\beta}_k$ are vectors that represent the output values and the net values for the units in the $k^\text{th}$ hidden layer.
As part of the error term for the units in the $l^\text{th}$ layer, backpropagation calculates $\frac{\partial \beta_{0,c}}{\partial\alpha_1} \frac{\partial \alpha_1}{\partial \beta_1}\dots\frac{\partial \alpha_l}{\partial \beta_l}$ as the error term associated with each network unit.
Thus, the only additional calculation for $h_i$ is:
$$\frac{\partial \boldsymbol{\beta}_{l}}{\partial v_{ri}} = \frac{\partial}{\partial v_{ri}} \sum_j \sum_t w_{jt}\: v_{rt}.$$
As before, $\frac{\partial \boldsymbol{\beta}_{l}}{\partial v_{ri}}$ is non-zero only when $t$ equals $i$.
For networks with at least one hidden layer:
\begin{equation}
	h_i=-\sum_j w_{ij} \delta_j.
	\label{eq_with_hidden}
\end{equation}
Equation~\ref{eq_with_hidden} is a strict generalization of Equation~\ref{eq_no_hidden}.
Equation~\ref{eq_no_hidden} only considers the one output unit, $c$, for which a known target value is being presented, whereas
Equation~\ref{eq_with_hidden} sums over each unit, $j$, into which the intrinsic value $v_{ri}$ feeds.

\subsection{Three-Phase Training}
To integrate generative backpropagation into the training process, LNN uses three phases to train $\mathbf{V}$ and $\mathbf{W}$:
1) the first phase computes an initial estimate for the intrinsic vectors, $\mathbf{V}$,
2) the second phase computes an initial estimate for the network weights, $\mathbf{W}$, and
3) the third phase refines them both together.
All three phases train using stochastic gradient descent. 
In phase 1, the intrinsic vectors are induced while there are no hidden layers to form nonlinear separations among them.
Likewise, phase 2 gives the weights a chance to converge without having to train against moving inputs.
These two preprocessing phases initialize the system (consisting of both intrinsic vectors and weights) to a good initial starting point, such that gradient descent is more likely to find a local optimum of higher quality.
Empirical results comparing three-phase and single-phase training show that three-phase training produces more accurate results than single-phase training, which only refines $\mathbf{V}$ and $\mathbf{W}$ together (see \cite{Gashler2014}).

\begin{algorithm}[tb]
	\caption{LNN($\mathbf{A},\mathbf{X}, \eta', \eta'', \gamma, \lambda$)}
	\label{alg_ubp}
	\begin{algorithmic}[1]
		\STATE Initialize each element in $\mathbf{V}$ with small random values
		\STATE Let $\mathbf{T}$ be the weights of a single-layer perceptron
		\STATE Initialize each element in $\mathbf{T}$ with small random values
		\STATE $\eta \leftarrow \eta'$; $s'\leftarrow\infty$
		\WHILE {$\eta > \eta''$}
			\STATE $s\leftarrow$ train\_epoch($\mathbf{A}$,$\mathbf{X},\mathbf{T},\lambda,\textbf{true},0$)
			\STATE {\bf if} $1-s/s'<\gamma$ {\bf then} $\eta \leftarrow \eta / 2$
			\STATE $s'\leftarrow s$
		\ENDWHILE
		\STATE Let $\mathbf{W}$ be the weights of a multi-layer perceptron with $l$ hidden layers, $l \ge 0$
		\STATE Initialize each element in $\mathbf{W}$ with small random values
		\STATE $\eta \leftarrow \eta'$; $s'\leftarrow\infty$
		\WHILE {$\eta > \eta''$}
			\STATE $s\leftarrow$ train\_epoch($\mathbf{A}$,$\mathbf{X},\mathbf{W},\lambda,\textbf{false},l$)
			\STATE {\bf if} $1-s/s'<\gamma$ {\bf then} $\eta \leftarrow \eta / 2$
			\STATE $s'\leftarrow s$
		\ENDWHILE
		\STATE $\eta \leftarrow \eta'$; $s'\leftarrow\infty$
		\WHILE {$\eta > \eta''$}
			\STATE $s\leftarrow$ train\_epoch($\mathbf{A}$,$\mathbf{X},\mathbf{W},0,\textbf{true},l$)
			\STATE {\bf if} $1-s/s'<\gamma$ {\bf then} $\eta \leftarrow \eta / 2$
			\STATE $s'\leftarrow s$
		\ENDWHILE
		\STATE {\bf return} $\{\mathbf{V},\mathbf{W}\}$
	\end{algorithmic}
\end{algorithm}

Pseudo-code for the LNN algorithm, which trains $\mathbf{V}$ and $\mathbf{W}$ in three phases, is given in Algorithm~\ref{alg_ubp}.
LNN calls the train\_epoch function (shown in Algorithm~\ref{alg_refine}) which performs a single epoch of training.
A detailed description of LNN follows.

Matrices containing the known data values, $\mathbf{X}$, and the item descriptions, $\mathbf{A}$, are passed into LNN along with the parameters $\eta', \eta'', \gamma, \lambda$ (defined below).
LNN returns $\mathbf{V}$ and $\mathbf{W}$.
$\mathbf{W}$ is a set or ragged matrix containing weight values for an MLP that maps from each $\mathbf{v}_i$ to an approximation of $\mathbf{x}_i\in\mathbf{X}$.

{\bf Lines~1-9} perform the first phase of training, which computes an initial estimate for $\mathbf{V}$.
{\bf Lines~1-4} initialize the model variables.
$\mathbf{T}$ represents the weights of a single-layer perceptron and the elements in $\mathbf{T}$ and $\mathbf{V}$ are initialized with small random values.
Our implementation draws values from a Normal distribution with a mean of 0 and a deviation of 0.01.
The single-layer perceptron is a temporary model that is only used in phase 1 to for the initial training of $\mathbf{V}$.
$\eta$ is the learning rate and $s'$ is used to store the previous error score.
As no error has been measured yet, it is initialized to $\infty$.
%
%
%
{\bf Lines~5-9} train $\mathbf{V}$ and $\mathbf{T}$ until convergence is detected.
$\mathbf{T}$ may then be discarded.
We note that many techniques could be used to detect convergence.
Our implementation 
decays the learning rate whenever predictions fail to improve by a sufficient amount.
Convergence is detected when the learning rate $\eta$ falls below $\eta''$.
$\gamma$ specifies the amount of improvement that is expected after each epoch, or else the learning rate is decayed.
$\lambda$ is the regularization term used in train\_epoch.

{\bf Lines~10-17} perform the second phase of training.
This phase differs from the first phase in two ways: 1) a multilayer perceptron is used instead of a temporary single-layer perceptron, and 2) $\mathbf{V}$ is held constant during this phase.

{\bf Lines~18-23} perform the third phase of training.
In this phase, the same multilayer perceptron that is used in phase 2 is used again, but $\mathbf{V}$ and $\mathbf{W}$ are both refined together.
Also, no regularization is used in the third phase.

\subsection{Stochastic gradient descent}

\begin{algorithm}[tb]
	\caption{train\_epoch($\mathbf{A}$,$\mathbf{X},\mathbf{W},\lambda,p,l$)}
	\label{alg_refine}
	\begin{algorithmic}[1]
		\FOR {{\bf each} known $x_{rc}\in\mathbf{X}$ in random order}
			\STATE $\mathbf{q}_r \leftarrow (\mathbf{v}_r, \mathbf{a}_r)$
			\STATE Compute $\alpha_c$ by forward-propagating $\mathbf{q}_r$ into a multilayer perceptron with weights $\mathbf{W}$
			\STATE $\delta_c\leftarrow (x_{rc}-\alpha_c)f'(\beta_c)$
			\FOR {{\bf each} hidden unit $i$ feeding into output unit $c$}
				\STATE $\delta_i\leftarrow w_{ic} \delta_c f'(\beta_i)$
			\ENDFOR
			\FOR {{\bf each} hidden unit $j$ in an earlier hidden layer (in backward order)}
				\STATE $\delta_j\leftarrow \sum_k w_{jk} \delta_k f'(\beta_j)$
			\ENDFOR
			\FOR {{\bf each} $w_{ij}\in\mathbf{W}$}
				\STATE $g_{ij}\leftarrow -\delta_j \alpha_i$
			\ENDFOR
			\STATE $\mathbf{W} \leftarrow \mathbf{W} - \eta (\mathbf{g} + \lambda\mathbf{W})$
			\IF {$p=\textbf{true}$}
				\FOR {$i$ {\bf from} $0$ {\bf to} $t-1$}
					\STATE {\bf if} $l=0$ {\bf then} $h_i\leftarrow -w_{ic} \delta_c$\\ {\bf else} $h_i\leftarrow -\sum_j w_{ij} \delta_j$
				\ENDFOR
				\STATE $\mathbf{v}_r \leftarrow \mathbf{v}_r - \eta (\mathbf{h} + \lambda\mathbf{v}_r)$
			\ENDIF
		\ENDFOR
		\STATE $s\leftarrow$ measure RMSE with $\mathbf{X}$
		\STATE {\bf return} $s$
	\end{algorithmic}
\end{algorithm}

For completeness, we describe train\_epoch given in Algorithm~\ref{alg_refine}, which performs a single epoch of training by stochastic gradient descent.
This algorithm is very similar to an epoch of traditional backpropagation, except that it presents each element individually, instead of presenting each vector, and it conditionally refines the latent variables, $\mathbf{V}$, as well as the weights, $\mathbf{W}$.

{\bf Line~1} presents each known element $x_{rc}\in\mathbf{X}$ in random order.
{\bf Line~2} concatenates $\mathbf{v}_r$ with the corresponding item description $\mathbf{a}_r$.
{\bf Line~3} computes a predicted value for the presented element given the current $\mathbf{v}_r$.
Note that efficient implementations of line 3 should only propagate values into output unit $r$.
{\bf Lines~4-10} compute an error term for output unit $r$, and each hidden unit in the network.
The activation of the other output units is not computed, so the error on those units is 0.
{\bf Lines~11-14} refine $\mathbf{W}$ by gradient descent.
{\bf Line 15} specifies that $\mathbf{V}$ should only be refined during phases 1 and 3.
{\bf Lines~16-19} refine $\mathbf{V}$ by gradient descent.
{\bf Line~22} computes the root-mean-squared-error of the MLP for each known element in $\mathbf{X}$.



\section{Experimental Results}
\label{section:results}
In this section we present the results from our experiments.
We examine LNN using the MovieLens\footnote{\url{http://www.grouplens.org}} data set from the HetRec2011 workshop \cite{RecSys2011}.
We use this data set because it provides descriptions for the movies in addition to the ratings matrix.
There are few data sets that provide user/item descriptions in addition to the ratings matrix (e.g. the Netflix data only contains user ratings).
Some data sets provide unstructured data such as twitter information or a set of friends on last.fm from which input variables could be created.
As this paper focuses on the performance of LNN rather than feature creation from unstructured data, we chose to use the MovieLens data set.
Also, running state-of-the-art recommendation systems can take a long time -- it was reported that running Bayesian probabilistic MF took 188 hours on the Netflix data \cite{Salakhutdinov2008}.
Using a smaller data set allows for a more extensive evaluation and facilitates cross-validation.
The MovieLens data set contains 2113 users and 10197 movies with 855598 ratings.
On average, there are 405 ratings per user and 84 ratings per movie.
For item descriptions, we use the genre(s) of the movie as a set of binary variables indicating if a movie belongs to one of the 19 genres.

We use LNN with and without three phase training.
This is equivalent to a hybrid UBP and hybrid NLPCA technique.
LNN with three phase training is denoted as LNN$_\text{3PT}$.
We compare LNN with several other recommendation systems:
1) content-boosted collaborative filtering (CBCF),
2) content-based filtering (CBF),
3) nonlinear principle component analysis (NLPCA),
4) unsupervised backpropagation (UBP), and
5) matrix factorization (MF).
For each recommendation system, we test several parameter settings.
CBF uses a single learning algorithm to learn the rating preferences of a user.
We experiment using na\"{i}ve Bayes (as is commonly used \cite{Melville2004}), linear regression, a decision tree, and a neural network trained with backpropagation.
The same learning algorithms are also used for CBCF and the number of neighbors ranges from 1 to 64.
For MF, the number of latent variables ranges from 2 to 32 and the regularization term from 0.001 to 0.1.
In addition to the values used for MF for the number of latent variables and the regularization term, the number of nodes in the hidden layer ranges from 0 to 32 for UBP, NLPCA, LNN, and LNN$_\text{3PT}$.
For each experiment, we randomly select 20\% of the ratings as a test set.
We then using 10\% of the training set as a validation set for parameter selection.
Using the selected parameters, we test on the test set and using 10-fold cross-validation.


\subsection{Results}
The results comparing LNN with the other recommendation approaches are shown in Table \ref{table:results}.
We report the mean absolute error (MAE) for each approach.
The bold values represent the lowest means within 0.002.
The algorithms that use latent variables are significantly lower than those that do not (CBCF and CBF), thus demonstrating the predictive power of using latent variables for item recommendation.
Latent inputs also allows one to bypass feature engineering -- often a difficult process.

\begin{table}[t]
\caption{The MAE from the investigated recommendation systems on the validation set and the test set.}
\label{table:results}
\centering
\begin{tabular}{l|cccc|ccc}
 & CBCF & CBF & LNN & LNN$_\text{3PT}$ & MF & NLPCA & UBP \\
\hline
Validation & 0.7709 & 0.8781 & \textbf{0.5885} & \textbf{0.5877} & \textbf{0.5886} & 0.6058 & 0.5942\\
Test & 0.7767 & 0.8831 & \textbf{0.5795} & 0.5810 & \textbf{0.5779} & 0.5971 & 0.5942 \\
10CV & 0.7754 & 0.8695 & \textbf{0.5781} & \textbf{0.5778} & \textbf{0.5760} & 0.5915 & 0.5915 \\
\end{tabular}

\end{table}

The addition of the item descriptions to NLPCA and UBP (LNN and LNN$_\text{3PT}$) improves the performance compared to only using the latent variables.
The performance of LNN and LNN$_\text{3PT}$ is similar to matrix factorization, which is widely considered state-of-the-art in recommendation systems when comparing MAE.
The power of LNN and LNN$_\text{3PT}$ comes when faced with the cold-start problem which we address in the following section.
As was discussed previously, MF and other pure collaborative filtering techniques are not able to address the cold-start problem despite being able to perform very well on items that have been rated previously a certain number of times.
(They also suffer from the gray sheep problem which occurs when an item has only been rated a small number of times.)
LNN and LNN$_\text{3PT}$ are capable of addressing the cold-start problem \textit{while} still obtaining similar performance to matrix factorization.


%

\subsection{Cold Start Problem}
To examine the cold-start problem, we remove the ratings for the top 10 most rated movies individually and collectively.
The number of removed ratings for a single movie ranged from 1263 to 1670 and 15,131 ratings were removed for all top 10.
The recommendation systems were trained using the remaining ratings using the parameter setting found in the previous experiments.
For LNN, predicting a new item poses an additional challenge since the latent variables for the new items have not been induced. 
We find that using 0 values for the latent inputs often produced worse results than CBF.
A CBF creates a model for each user based on item descriptions and corresponding user ratings.
LNN, on the other hand, produces a single model which is beneficial when using all of the ratings because the mutual information between users and items can be shared.
The shared information is contained in the latent variables.
The quality of the latent variables depends on the number of ratings that a user has given and/or an item has received.

\begin{algorithm}[tb]
	\caption{new\_item\_prediction($\mathbf{a}_{newItem}$)}
	\label{alg_prediction}
	\begin{algorithmic}[1]
		\STATE Let $count$ be a map containing the count of how many times each rating was predicted
		\STATE Initialize each element in $count$ to 0
               \STATE $numNeighbors \leftarrow 100$; $distThresh \leftarrow 0$
		\STATE $neighbors \leftarrow$ getNeighbors($\mathbf{a}_{newItem}$, $numNeighbors$)
		\FOR {$i$ {\bf from} 0 {\bf to} $numNeighbors - 1$}
			\STATE $numRatings \leftarrow$ count number of ratings for $neighbors[i]$
			\IF {$numRatings > 50$ \&\& $distance(neighbors[i]) > distsThresh$}
				\STATE $\mathbf{q}_{new} \leftarrow (\mathbf{v}_{neighbors[i]}, \mathbf{a}_{newItem})$
				\STATE $prediction \leftarrow $ rounded prediction of $\mathbf{q}_{new}$
				\STATE $counts[prediction] += numRatings$
			\ENDIF
		\ENDFOR
		\STATE {\bf return} maxIndex($counts$)
	\end{algorithmic}
\end{algorithm}

To compensate for the lack of latent variables for the new items, we utilize the new\_item\_prediction function that takes a vector $\mathbf{a}_{newItem}$ representing the description of the new item and is outlined in Algorithm 3.
At a high level, new\_item\_prediction uses $\mathbf{a}_{newItem}$ to find its nearest neighbors.
The induced latent input variables for each neighbor are concatenated with $\mathbf{a}_{newItem}$ and fed into a trained LNN to predict a rating for the new item.
The weighted mode of the predicted ratings of the new item is then returned.
The rating from each neighbor is weighted according to how many times it has been rated.
By weighting, we mean when selecting the mode from a set of numbers, the predicted rating is added $r$ times to the set where $r$ is the number times that the neighbor item has been rated.
We chose to use the mode rather than the mean because the mode is more robust to outliers and achieves better empirical results on the validation sets in our experimentation.
We next describe new\_item\_prediction in more detail.

{\bf Lines~1-2} initializes a counter that keeps track of how many times a rating has been predicted for the new item and initializes all values to 0.
{\bf Line~3} initializes the number of nearest neighbors to search for to 100 and sets the distance threshold to 0.
We chose 100 neighbors because it was generally more than enough neighbors to produce good results.
As we used binary item descriptions of movie genres, we only considered using the latent variables from items that have the same genre(s) (has a distance of 0).
These values come into play in {\bf line 7} where an item is not used if its distance is greater than $distThresh$ (in this case 0), and if an item has not been rated at least 50 times.
The value of 50 was chosen based on the evaluation of a content-based predictor \cite{Mitchell1997ML}.
The number of times that an item has been rated helps to determine the quality of the induced latent variables for an item and provides a confidence level for latent variables.
{\bf Line~4} finds the closest neighbors and inserts their indexes into an array.
{\bf Lines~5-10} count the number of times that each rating is predicted weighted by the number of times that the item has been rated.
We use a linear rating such that the prediction for an item that has been rated 100 times will count for 100 ratings of that predicted value.
This helps to discount items that have only been rated a few times and whose latent variables may not be set to good values.
{\bf Line~13} returns the index (rating) that has the max count (i.e. the mode).

The results for recommending new items using new\_item\_prediction are provided in Table \ref{table:newItems}.
The values at the top of the table correspond to the movie id in the MovieLens data set.
The bold values represent the lowest MAE value obtained.
No single recommendation system produces the lowest MAE all of the items, suggesting that some recommendation systems are better than others for a given user and/or item as has been suggested previously \cite{Lee2012}.
LNN and LNN$_\text{3PT}$ each score the lowest MAE for several movies individually.
With the exception of movie 2571, LNN and LNN$_\text{3PT}$ produce the lowest MAE for all of the movies when they have not been previously rated.
When holding out all 10 items, LNN$_\text{3PT}$ produces the lowest MAE value.
This shows the importance of using latent variables.
CBCF uses CBF to create a dense matrix (except for the ratings corresponding to the active user) and then uses a collaborative filtering technique on the dense matrix to recommend items to the user.
Thus, more emphasis is given to the CBF which generally produces poorer item recommendations than a collaborative filtering approach.
LNN, on the other hand, utilizes the latent variables and their predictive power.

\begin{table}[t]
\caption{The MAE for the top 10 most rated movies (individually and combined) when held out of the training set.}
\label{table:newItems}
\centering
\setlength{\tabcolsep}{4.65pt}
\begin{tabular}{l|cccccccccc|c}
alg & 2571 & 2858 & 2959 & 296 & 318 & 356 & 480 & 4993 & 5952 & 7153 & top10 \\
\hline
CBCF & \textbf{0.889} & 0.898 & 0.875 & 0.742 & 0.929 & 0.760 & 0.720 & 0.755 & 1.053 & 0.981 & 0.896 \\
CBF & 0.957 & 0.905 & 0.920 & 0.870 & 0.965 & 0.866 & 0.766 & 0.790 & 1.121 & 1.041 & 0.972 \\
LNN & 1.175 & \textbf{0.689} & 0.894 & \textbf{0.666} & \textbf{0.789} & \textbf{0.593} & 0.552 & \textbf{0.558} & 0.577 & \textbf{0.523} & 0.859 \\
LNN$_\text{3PT}$ & 1.189 & \textbf{0.690} & 0.906 & 0.680 & 0.810 & \textbf{0.595} & \textbf{0.541} & 0.587 & \textbf{0.566} & \textbf{0.521} & \textbf{0.847} \\
\end{tabular}
\end{table}

\subsection{Efficiency}
The efficiency of LNN is not precise as is the case for most neural network models since it is based on the number of iterations until convergence.
In our experiments, LNN always converges regardless of the parameter settings.
However, some parameter settings did require longer to reach convergence than others.
The average time in seconds required to run each algorithm using the parameter settings found in the previous experiments is shown in Table \ref{table:time}.
The additional complexity of LNN requires more time to train.
However, it has the benefit that a new model will not have to be induced in order recommend new or unrated items as is the case with MF, NLPCA, and UBP.
For recommending new items in LNN, LNN uses a k-d tree for the nearest neighbor search which has $log(n)$ search and insert complexities.

\begin{table}[t]
\caption{The time (in seconds) taken to run each algorithm.}
\label{table:time}
\centering
\begin{tabular}{l| ccccccc}
 & CBCF & CBF & LNN & LNN$_\text{3PT}$ & MF & NLPCA & UBP \\
\hline
train & 2278.2 & 9.1 & 43.4 & 60.2 & 4.8 & 5.8 & 5.8 \\
Ave 10CV & 2432.7 & 9.6 & 53.9 & 193.4 & 7.6 & 8.5 & 10.0 \\
\end{tabular}
\end{table}

\section{Conclusions and Future Work}
In this paper, we presented a neural network with latent input variables capable of recommending unrated items to users or items to new users which we call a \textit{latent neural network} (LNN).
The latent variables and input variables allow information and correlations among the rated items to be represented while also incorporating the item descriptions in the recommendation.
Thus, LNN is a hybrid recommendation algorithm that leverages the advantages of collaborative filtering and content based filtering.

Empirically, a LNN is able to achieve similar results to state-of-the-art collaborative filtering techniques such as matrix factorization while \textit{also} addressing the cold-start problem.
Compared with other hybrid filters and content-based filtering, LNN achieves much lower error when recommending previously unrated items.
As LNN achieves similar error rates to the state-of-the-art filtering techniques and can make recommendations for previously unrated items, LNN does not \textit{have} to be retrained once new items are rated in order to recommend them.

As LNN is built on a neural network, it is capable of modeling higher-order dependencies and non-linearities in the data.
However, the data in the MovieLens data set and many similar data sets is well suited to using linear models such as matrix factorization.
This may be due in part to the fact many of the data sets are inherently sparse and nonlinear models could overfit them and reduce their generalization.
As a direction of future work, we are examining how to better incorporate the non-linear component of LNN.
%
We are also looking at integrating \textit{both} user and item descriptions with latent input variables to address the new user problem \textit{and} the new item problem in a single model.

\bibliographystyle{abbrv}
\begin{small}
\bibliography{/home/msmith/Research/bibliography}
\end{small}

%
%
%
%

\end{document}